\documentclass[10pt, a4paper]{article}
\usepackage{lrec-coling2024} % this is the new style
\usepackage{xcolor}
\usepackage{hyperref}
 \definecolor{darkblue}{rgb}{0, 0, 0.5}
  \hypersetup{colorlinks=true, citecolor=darkblue, linkcolor=darkblue, urlcolor=darkblue}

\usepackage{xstring}

\usepackage{color}

\title{Detecting Sexual Content at the Sentence Level \\in First Millennium Latin Texts}

\name{Thibault Clérice} 

\address{Inria \\ Paris, France \\
         \{name.surname\}@inria.fr\\}

\abstract{In this study, we propose to evaluate the use of deep learning methods for semantic classification at the sentence level to accelerate the process of corpus building in the field of humanities and linguistics, a traditional and time-consuming task. We introduce a novel corpus comprising around 2500 sentences spanning from 300 BCE to 900 CE including sexual semantics (medical, \textit{erotica}, etc.). We evaluate various sentence classification approaches and different input embedding layers, and show that all consistently outperform simple token-based searches. We explore the integration of idiolectal and sociolectal metadata embeddings (centuries, author, type of writing), but find that it leads to overfitting. Our results demonstrate the effectiveness of this approach, achieving high precision and true positive rates (TPR) of respectively 70.60\% and 86.33\% using Hierarchical Attention Networks (HAN). We evaluate the impact of the dataset size on the model performances (420 instead of 2013 training samples), and show that, while our models perform worse, they still offer a high enough precision and TPR, even without masked language models (MLM), respectively 69\% and 51\%. Given the result, we provide an analysis of the attention mechanism as a supporting added value for humanists in order to produce more data. 
 \\ \newline \Keywords{Latin, sentence classification, figurative speech, sexuality, low-resource} }

\begin{document}

\maketitleabstract

% \textit{\color{purple} Warning: this paper contains sexual content that may be offensive. Original Latin text is given in the main body of the text, English translations and comments (when required) are provided at the end of the document.}

\section{Introduction}

% What is the problem you want to solve

Classical and medieval studies encompass the history of over two millennia of human activities. A traditional approach to research in these fields involves constructing a corpus centered around specific themes, such as colors \citep{goldman2013color}, and subsequently analyzing the corpus to draw historical or philological conclusions. To build such a corpus, researchers rely on three main sources: (1) their own knowledge of the subject and the corpus of the centuries they study, (2) accumulated knowledge from previous researchers working on similar topics, and (3) plain-text search on closed or open corpora. While this secular approach has proven effective, it faces limitations under certain conditions due to (a) the inherent vagueness of the subject, (b) its lexical diversity, or (c) its temporal coverage. For instance, examining the topic of sexualities in Latin from antiquity (300 BCE) to the high Middle Ages (900 CE) presents all three challenges: (i) sexuality encompasses reproductive aspects (including manifestations in medical texts), recreational aspects (found in poetic texts), and social aspects (appearing in political or historical texts); it encompasses both human and animal sexuality, and the connections between lexical usages within this domain are significant; (ii) furthermore, sexuality in ancient Rome, much like in modern times, is a lexically rich field, featuring unique metaphors that require contextual understanding; (iii) finally, the corpus for this period amounts to 120 million words, exceeding human memory capacity.

Three types of topic manifestations can be distinguished: 
\begin{itemize}
    \item specific lexemes within the semantic field, e.g. ``Quid faciat uolt scire Lyris. Quod sobria: fellat.\footnote{``Lyris wants to know what she does [drunk]. What she does sober: she sucks (\textit{fellat}).''}'', Martial, \textit{Epigrammata}, II, 73. 
    \item lexemes used metaphorically within the field, e.g. ``Tanta est quae Titio columna pendet , Quantam Lampsaciae colunt puellae.\footnote{``The column (\textit{columna}) that hangs from Titius is so large that the young girls of Lampsaka venerate it.''}'', Martial, \textit{Epigrammata}, XI, 51.
    \item complex and unique figurative speech, e.g. ``Donec proterua nil mei manu carpes, licebit ipsa sis pudicior Vesta.\footnote{``As long as you don't pick anything from me with a shameless hand, you'll be chaster than Vesta herself.'' (Implying that if someone steals something from Priapus, who is talking here, they will be sexually assaulted).}'',  Anonymous, \textit{Priapea}, 31.
\end{itemize}

These manifestations pose challenges for corpus construction and analysis. While the first category is straightforward to include, the second requires filtering, and the third cannot easily be retrieved through plain text searches. Addressing these challenges is crucial for studying subjects like sexualities, which involve intricate figurative language.

% How does the literature touch such a problem ?
Although there is no existing research specifically focused on sentence classification in Latin regarding specific semantic features, studies have been conducted on English corpora, particularly in the context of sexual content classification or the detection of sexual abusers, as well as metaphor detection at the token level. Most recent approaches in this domain predominantly employ recurrent neural networks (RNN) or masked language models (MLM) for classification tasks.

Latin presents specific challenges as a lower-resource language compared to English due to its extensive historical existence spanning over two millennia and its highly rich morphology. While Latin BERT \citep{bamman2020latin} has been developed, its application to sentence classification remains unexplored. Previous Latin classification tasks primarily focus on text (specifically for authorship attribution) or token-level analysis (specifically lemmatization or named entity recognition), lacking specific research on sentence classification for semantic features. These challenges highlight the need for evaluating different sentence encoders in Latin natural language processing (NLP).

We address this as a binary sentence classification problem, where inputs are the original tokens of the sentence as a sequence, and output class are either positive (for sexual semantics) or negative.

In this paper, we propose to assess the effectiveness of various sentence encoders (Long Short-Term Memory [LSTM], Gated Recurrent Units [GRU], sentence-level HAN, and BERT pooling) and input (lexemes or lemma) for our task. Our approach includes the evaluation of  relevant categorical features \citep{kim_categorical_2019} related to the text (verse/prosaic, work structure) and the author (century of birth, identity). The inclusion of these features aims to reduce the weight of the rich morphology, while also addressing idiolectal or sociolectal specificities.

We constructed a corpus consisting of 2,516 sample sentences focused on sexual content, including figurative speech (metaphors, comparisons, etc.). This corpus allowed us to evaluate the effectiveness of various sentence encoders and embeddings for semantic classification.

Our main contributions can be summarised as follows:
\begin{itemize}
    \item Introduction of a new dataset (\textit{Lasciva Roma Dataset}) for a new task (Sentence classification) for Latin.
    \item Experiments and evaluation of the best architectures for expanding such dataset depending on the dataset size.
    \item Evaluation and rejection of the use of categorical features identifying idiolectal or sociolectal features of text in an imbalanced dataset.
    \item An analysis of the usefulness of attention mechanism and its relation to the original annotations in the context of filtering positive matches.
    \item Freely distributed scripts and models.
\end{itemize}

The remainder of the paper is organized as follows: subsection \ref{sec:bg} provides background information on current approaches and the inherent difficulties associated with Latin NLP, followed by a brief introduction to similar semantic tasks for English language (metaphor identification, sexual content identification). Section \ref{sec:method} details the different architectures and embeddings employed for categorical and morpho-syntactic features. Section \ref{sec:data} presents the corpus, its coverage and its biases. Section \ref{sec:experiments} presents experiments and their set-up. In section \ref{sec:results}, we discuss the results, including the bias testing of categorical feature embeddings and an in-depth analysis of the attention in HAN networks.

\section{Background and Related Work}
\label{sec:bg}

\subsection{Latin, Ancient Greek and Natural Language Processing}
Classical studies have historically shown an interest in computational approaches, with the first stylometric approach by researchers on Plato dating back to the 19th century \citep{lutoslawski1898principes}, and digital concordances on punch cards in partnership with IBM as early as 1949 \citep{busa1980annals}. Latin and Ancient Greek (AG), the two most extensively studied languages in the Western tradition of classical studies, encompass a wide chronological range (Latin: 3 c. BCE--present day) and a vast geographical scope, while also exhibiting a highly complex morphological system. These three characteristics greatly impact our ability to computationally analyze these languages, as they continue to be considered low-resource languages in terms of NLP. Within this context, three distinct tasks emerge in NLP related to classical studies: lemmatization, authorship attribution and verification, and intertextuality detection.

Lemmatization and morphosyntactic tagging in Latin and AG have posed longstanding challenges, dating as far back as 1962 when researchers from the \textit{Laboratoire d'Analyse Statistique des Langues Anciennes} (LASLA) started envisioning automatic taggers \citep{delatte_programme_1965}. Early research up until the 1990s primarily focused on rule-based annotation without considering contextual factors. Statistical models such as TreeTagger \citep{schmid1994treetagger} have impacted computational classics by reducing uncertainties in tagged corpora. With the advent of RNNs, works such as \citet{manjavacas_improving_2019} in combination with large corpus such as LASLA data (around 1.7 million tokens \citealp{longree2010structures}) have provided good generalizing lemmatizers. Lemmatization and morpho-syntactic tagging of Latin have recently garnered significant interest within the framework of events such as the "Ancient Language Processing Workshop" \cite{anderson2023proceedings}, the EvaLatin competition \cite{sprugnoli2022overview} held in conjunction with the "Workshop on Language Technologies for Historical and Ancient Languages"~\cite{sprugnoli2022proceedings}, and the LiLa ERC project \cite{passarotti2019lila}. However, despite the existence of several datasets for Latin lemmatization and morpho-syntactic tagging, specifically in Universal Dependencies repositories~\cite{de2021universal}, they have shown potentially divergent approaches to lexicalization based on their underlying lexicon \cite{cleri2022} or their methodology for morpho-syntactic tagging \cite{gamba2023latin}, thereby diminishing their interoperability.

Authorship analysis in classical philology has long been a tradition, involving the search for authority and the identification of authors in anonymous, modified, or composite texts. Authorship attribution has experienced a resurgence in the digital humanities, employing lexical \citep{kestemont_authenticating_2016}, syntactic \citep{gorman_author_2020}, and metrical features \citep{nagy_metre_2021}. These mostly employs both unsupervised approaches (hierarchical clustering and distance-based metrics), and supervised approaches, particularly with SVM.

Quantitative and computational studies have revitalized the field of intertextuality \citep{forstall_quantitative_2019}, which examines textual transfers within discourses. In classical philology, intertextuality involves tracing connections between texts, including citations, allusions, and reworked passages. Detecting intertextual elements, such as clichés and stereotypes, is accomplished through methods like fuzzy matching \citep{coffee_tesserae_2013}, similarity measurements \citep{forstall_quantitative_2019}, and more complex neural networks \citep{manjavacas_feasibility_2019}. Challenges arise from the vast number of potential sources, the complexity of detecting subtle allusions, and the transition from lexical to semantic intertextuality.

Content classification outside of authorship analysis remains a rare task in relation with Latin processing, but recent examples have shown that it remains a viable field. \citet{classifyinglatin} have used feature based classifier to sort a corpus of epigraphic documents and \citet{picca2023unveiling} have used a lexicon approach for emotion detection in 26 Latin plays. Using more advanced approaches, \citet{riemenschneider2023graecia} have used various forms of transformers to detect ``Latin allusions to Ancient Greek literature'' which showed great potential as a new step for intertextuality detection across languages.

\subsection{Metaphor identification}
Regarding linguistic metaphor identification \citep{steen_method_2010}, while closely aligned with the tradition established by \citet{lakoff_metaphors_2003}, the definition used focuses on detecting figurative language at the token level. The task primarily revolves around detecting the figurative sense of individual tokens rather than phrases. For example, the phrase "To emerge as the winner of a debate" would be seen as an accumulation of metaphors, with "winner" borrowing from the realm of combat and "emerge" conveying the idea of movement or departure.

The first \textit{Shared Task on Metaphor Detection} \citep{leong_report_2018} at \textit{North American Chapter of the Association for Computational Linguistics} (NAACL) witnessed the use of two out of three tools that employed not only semantic information carried by embeddings but also part-of-speech (POS) annotation. \citet{stemle_using_2018}'s word on ``bot.zen'' incorporated various embeddings, including some trained on corpora of individuals learning English, based on the assumption of an uneven use of figurative language between language learners and native speakers.

The second competition took place in 2020 \citep{leong_report_2020} during the annual \textit{Association for Computational Linguistics} (ACL) conference. This competition showcased the significant presence of transformer-based architectures (RoBERTa, BERT, AlBERT, etc.) in the field, with these architectures consistently occupying the top five positions out of twelve participants. The best-performing model, "DeepMet" utilized transformer-based embeddings and morphosyntactic information (specifically two forms of POS). Evaluation primarily relied on the F1-score, with the highest scores reaching 76.9\% across all morphosyntactic categories on the Vrije Universiteit Amsterdam dataset for metaphor detection (commonly known as the VUA metaphor dataset) \citep{steen_method_2010}. %Additionally, these competitions explored success rates and performance disparities among different genres (Academic, Fiction, Press, Conversations).% to gain deeper insights into what these algorithms can learn.

While metaphor identification provides us with intriguing avenues for exploration, particularly through the concatenation of different embeddings trained on diverse datasets, the semantics of "metaphor" in this context diverge significantly from its usage in stylistics, emphasizing single-word metaphoric sense detection over the detection of multi-word metaphors. 

\subsection{Sexual content identification}

The specific example of sexual content classification spans across various domains, from identifying sexual predators to classifying songs for commercial purposes. The PAN competition's ``Sexual Predator Identification'' dataset \citep{inches_overview_2012} remains the largest available dataset, addressing the implicit and discursive progression aspects of sexual predation. Similar to toxicity detection, the field has transitioned from Bag-of-Words models to deep learning approaches. Recent models, such as \citet{munoz_smartsec4cop_2020}'s, utilize Word2Vec projection and convolutional neural networks, while \citet{vogt_early_2021}'s approach incorporates BERT projection and temporal dimensions, achieving higher F1-Scores.

Regarding song lyrics classification, \citet{fell_comparing_2019} explore various methods, including dictionary attacks, Bag-of-Words, Convolutional Neural Networks (CNN), Hierarchical Attention Network (HAN, \citealp{yang_hierarchical_2016}) and BERT-based models. They show that the CNN model performs well in the explicit category, while BERT outperforms others in the non-explicit category, which dominates the corpus. % Grammatical information was not integrated into the network, despite the ambiguity of words like ``bitch'' (to bitch about something) and their potential reuse in non- or less-vulgar interpretations in songs\footnote{As an example, they specifically mention the reuse of the word by Meredith Brooks in her song ``I'm a bitch''.}.

\section{Proposed method}
\label{sec:method}

\begin{figure}[ht]
    \centering
    \includegraphics[width=\linewidth]{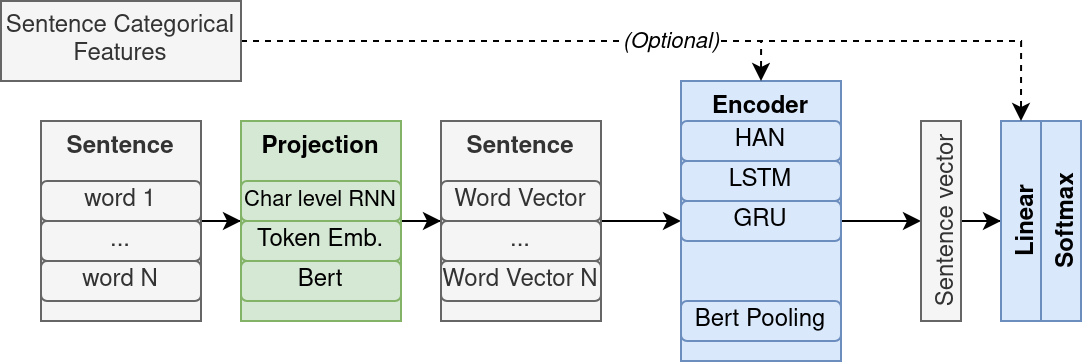}
    \caption{Structure of the general architecture. Sentence categorical features are used (a) in the encoder, (b) in the linear layer, (c) or not used.}
    \label{fig:layers}
\end{figure}

We propose to evaluate the variation of our system, which consists of two modules: the projection layer and the encoding layer (see Figure \ref{fig:layers}). The projection layer utilizes contextual or non-contextual embeddings to generate a vector for each token in the input. On the other hand, the encoding layers produce a single vector, which is then passed to a linear layer.

\subsection{Embedding layers}

\begin{figure}[ht]
    \centering
    \includegraphics[width=\linewidth]{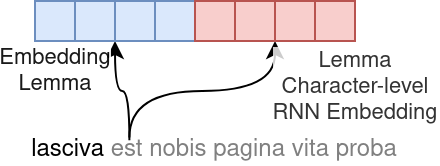}
    \caption{Concatenation of features to represent word-level informations.}
    \label{fig:embeddings}
\end{figure}

In the context of the embedding layer, we explore variations of commonly used approaches in similar token or sentence classification tasks. These include lexem embedding and lemma embedding, pretrained on the whole corpus of Latin texts~\citeplanguageresource{thibault_clerice_2020_corpus_latin}. Additionally, we employ character-level encoding using a BiLSTM encoder for lemmas or lexemes. Lemma are provided using the Pie-Extended LASLA tagger~\citep{manjavacas_improving_2019}\footnote{We opted for this tagger because it was trained on the most consistent dataset in terms of annotation practices, closely aligned with the timeframe of the dataset used in our paper. Specifically, the dataset comprises the harmonized LASLA dataset, the \textit{Vulgate} from the PROIEL dataset adapted to LASLA's annotation practices \citelanguageresource{haug2008creating} available at \url{https://github.com/PonteIneptique/proiel-to-lasla}, the Priapea corpus \citelanguageresource{Clerice_Lemmatisation_et_analyse_2021}, and excerpts from late antiquity Latin texts \citelanguageresource{Glaise_Du_IIeme_siecle_2021}. Pie-Extended was specifically selected because it accommodates the idiosyncrasies of the LASLA dataset, such as missing punctuation, initial capitalization, and other common features typically absent in texts from open repositories.}. Token level and character level information are concatenated (see Table \ref{tab:embedding}). %Furthermore, we experiment with the use of \cite{stroebel-roberta-base-latin-cased1}'s Latin RoBERTa Model as contextual embeddings. 

\begin{table}[ht]
\resizebox{\linewidth}{!}{%}
\begin{tabular}{lllrl}
\hline
Type of information      & Level     & Type           & Size                 & Pretraining   \\ \hline \hline
Lexem                    & Token     & Classic        & 200                  & Word2Vec      \\
Lexem                    & Character & BiLSTM         & 300                  &               \\
Lexem                    & Subword   & BERT           & 762                  & BERT, RoBERTa \\ \hline
Lemma                    & Token     & Classic        & 200                  & Word2Vec      \\
Lemma                    & Character & BiLSTM         & 300                  &               \\ \hline
Categorical              &           &                & 64                   &               \\ \hline
\end{tabular}
}
\caption{List of the embedded information explored in the evaluation. In the case of a model utilizing Lemma(Token) and Lemma(Character), the projection dimensionality is 500. For character-level encoding of lemma or token, the character embedding dimensionality is 100, and the BiLSTM outputs a vector of size 300.}
\label{tab:embedding}
\end{table}

\subsection{Encoding layers}

Regarding the encoding layers, we employ widely used approaches for sequence encoding, including RNN models such as BiLSTM and GRU. Additionally, we incorporate the sentence-level attention mechanism proposed by \citet{yang_hierarchical_2016} in their hierarchical attention network for documents. For BERT embeddings, we extend our encoding layer with pooling strategies such as mean reduction, maximum reduction, concatenation of both, and beginning of sequence [BOS] tokens (see Table \ref{tab:encoder}).

\begin{table}[ht]
\resizebox{\linewidth}{!}{%}
\begin{tabular}{lll}
\hline
BERT-Only & Encoder Type & Output dimension tested \\ \hline \hline
             & BiLSTM       & 128/256                 \\
             & GRU          & 128/256                 \\
             & HAN (LSTM)   & 128/256                 \\
Yes          & MeanMax      & 1536                    \\
Yes          & Mean         & 768                     \\
Yes          & Max          & 768                     \\
Yes          & {[}BOS{]}    & 768                     \\ \hline
\end{tabular}
}
\caption{List of encoders, with output dimension.}
\label{tab:encoder}
\end{table}

\subsection{Categorical feature usages}

``When metaphor is 'in absentia,' it establishes a symbolic connection that must be identified through concordant conjectures about discourse, the type of work, the genre of the text, and the idiolectal hierarchization of isotopies'' \citep[p. 98]{rastier_tropes_1994}. This quote from F.~Rastier emphasizes the significance of context in understanding figurative speech. Different genres and authors have distinct vocabularies and metaphorical spaces, which can be utilized to provide additional information to the network about the broader context (author, century, etc.).

The formalization of context and its integration into classification and sequential language processing is a relatively new area of research. One notable publication on this subject is the work by \citet{kim_categorical_2019}. The authors introduce information about the author in parallel to the text at various insertion points, such as Embedding, Attention, LSTM, and Linear layers. They project metadata into multiple embedding spaces and then integrate it through network modifications. In order to evaluate this approach, the authors used restaurant reviews with author identification provided as metadata. One of the given example revolve around the ambiguous sense of ``spicy'', which could be positive or negative depending on the reviewer. By incorporating information related to the reviewer, the authors demonstrate an increase in classification accuracy of up to +4.45 percentage points (F1).

In our network, we reuse their approach by extending this idea of incorporating idiolectical (author-level) to sociolectal (period, genre) features. We evaluate four different metadata features in their own embedding space: (1) Author identity; (2) Century of birth of the author; (3) Form of the text (verse or prose); (4) Editorial structure of the text, which can serve as a proxy for the genre (\textit{e.g.}, book/poem, book/chapter, letter, etc.).

\section{Data}
\label{sec:data}

\subsection{Overview}
\label{subsec:positive}

\paragraph{Positive examples}

We present a novel dataset based on the \textit{Latin sexual vocabulary} written by \citet{adams}, which provides a detailed "[description and classification of] the varieties of language used in Latin to refer to sexual parts of the body, sexual acts, and excretion" \citep[p.~2]{adams}. In this book, J.~N.~Adams provides series of phrases (single or multi-words) carrying sexual meaning. From this book, we produced a dataset of over 702 unique lemma -- for some used only in combination with others -- in 2,516 sentences describing sexual acts, genitals or other sexualized parts of the body. 

In his book, Adams provided either (1) clear references to examples, (2) references to lexicons which in turn contained references to passages in Latin literature or (3) just the lemma in case of a word whose sense is never ambiguous. We collected each sentence from an open corpus of Latin Litterature. Each example is automatically tagged (lemma, POS, and morpho-syntactical data [MSD]) using the Pie lemmatizer \citep{manjavacas_improving_2019} and a model based on LASLA data \citep{Clerice_Latin_Deucalion_a}.

\paragraph{Period biases}

\begin{figure}[ht]
    \centering
    \includegraphics[width=\linewidth]{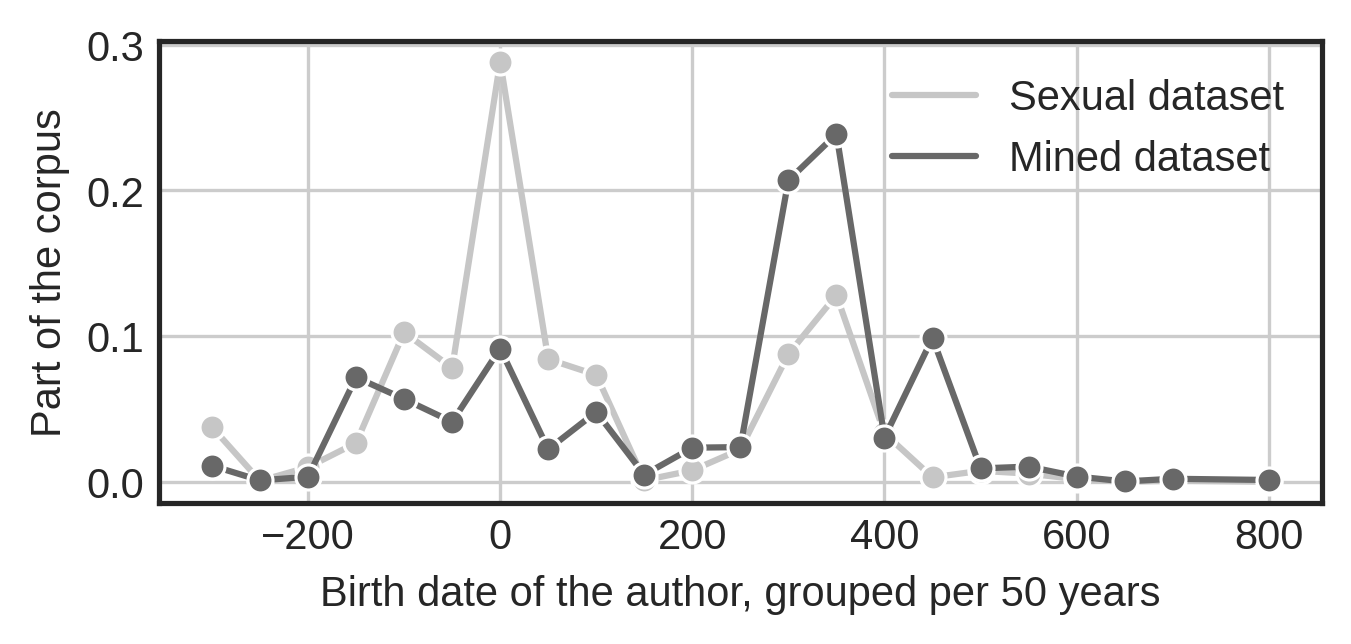}
    \caption{Percentage of the corpus (in number of words) per 50-years increment. }
    \label{fig:biases-centuries}
\end{figure}

The resulting dataset reflects not only the biases inherent in the representation of the subject matter in Latin literature (as discussed below) but also the biases of Adams himself towards the classical corpus up to the 3rd century CE (see Figure \ref{fig:biases-centuries}). This inclination towards "classical" Latin may potentially be attributed to the size of the available corpora for conducting evidence searches. While the first six centuries of Latin literature comprise only a few million words, the Late Antiquity and high Middle Ages (prior to the 9th century CE) encompass over 80 to 120 million words, depending on the sources considered. Given the technological limitations of the 1980s, it is understandable that the coverage of late Latin would be more partial.

\paragraph{Style biases}

\begin{figure}[ht]
    \centering
    \includegraphics[width=\linewidth]{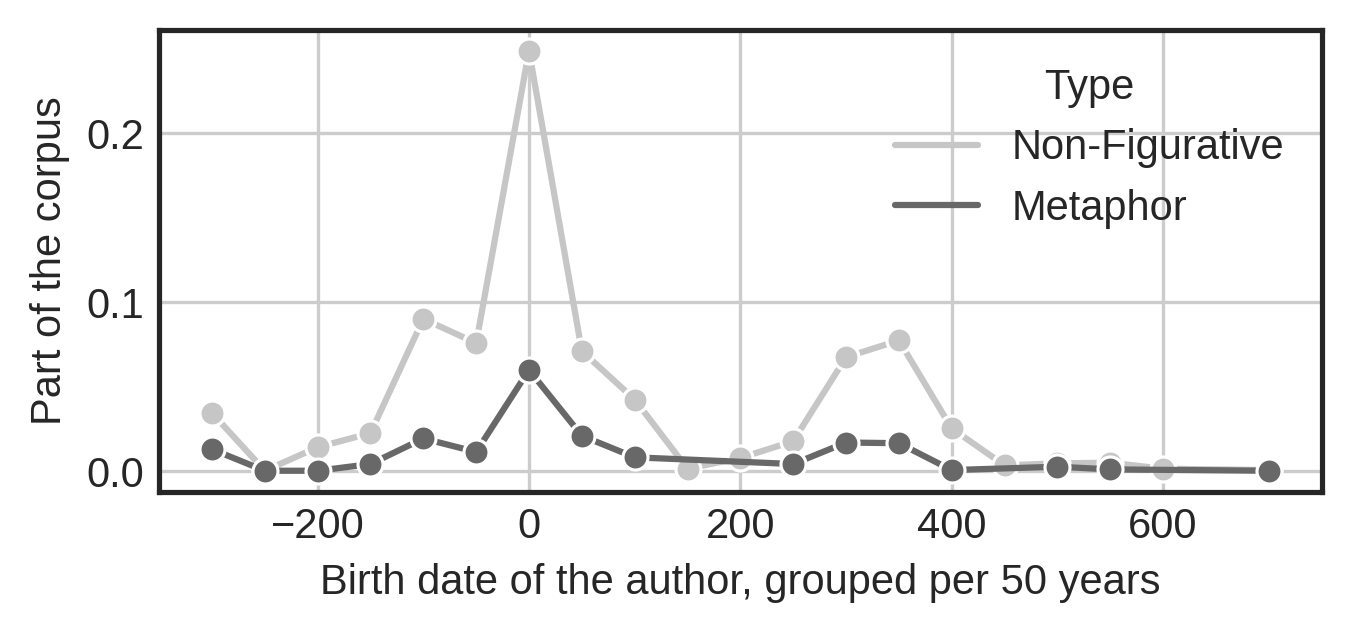}
    \caption{Percentage of the corpus (in number of examples) per 50 year increment regarding the type of examples. Metonimies are counted as non-figurative.}
    \label{fig:biases-content-type}
\end{figure}

We followed Adams' approach by annotating in each sentence the token(s) that bear the sexual sense, along with their "stylistic" features, specifically indicating whether the word was used metaphorically or metonymically. The resulting corpus exhibits a significant bias towards the 1st century CE across both categories but specifically for non-figurative language (see Figure \ref{fig:biases-content-type}). Two specific works contribute significantly to these biases: Martial's \textit{Epigrammata} and the anonymous \textit{Priapea}. The dataset contains approximately 60\% of the words from the \textit{Priapea}, while Martial's works have 350 excerpts included in the dataset, accounting for 10\% of its words (more than double the frequency of the second most represented work).

\paragraph{Negative examples}

Unlike the positive examples, sentences that do not contain sexual semantics can end up addressing any other subject. In order to deal with this randomness, we mined the lemmatized corpus of Latin we used previously for searching lemmas~\citelanguageresource{thibault_clerice_2020_corpus_latin}. For each work in the corpus (\textit{e.g.} Cicero's \textit{De Finibus}), we randomly draw 30 sentences. We check the sentences against our positive database, in the event that the sentence is a known positive. The resulting ``negative'' corpus amounts to 24,924 sentences.

\subsection{Corpus splits}
\label{subsec:splits}

In regards with our original goal, which is to support corpus creation in historical linguistics and digital humanities, we provide two different splits for our corpus:
\begin{enumerate}
    \item ``Full corpus'': A full, unfiltered corpus used to test the machine's ability to learn on the largest sample we had. This experiment involved approximately 2000 training samples and 252 evaluation samples (see Table \ref{tab:sample-size}).
    \item ``Partial corpus'': A randomly sampled corpus based on the full one, containing only 420 positive and 3970 negative training samples, but using the same dev (251 samples) and test samples (251 positive).
\end{enumerate}

While the full corpus remains a relatively small corpus in comparison with Yelp 2013's 80,000 samples~\citeplanguageresource{tang2015learning}, the time to produce 2250 examples (not counting test) remains high for a majority of projects. In order to evaluate the feasibility with a smaller dataset, we downsample the corpus to 672 samples (with a stable dev pool) which would be less expensive to produce.

The intention behind the partial corpus is to serve as a benchmark for assessing ``how much data is needed'' to develop a model, particularly for individuals aiming to construct similar tools within their specific lexical fields. Keeping the same test set allows for evaluating the impact of the downsizing of the train set.

\begin{table}[ht]
    \centering
    \resizebox{\linewidth}{!}{%}
        \begin{tabular}{lrrr|rrr}
        \hline
                       & \multicolumn{3}{c|}{Positive samples} & \multicolumn{3}{c}{Negative samples} \\ 
                       & Train       & Dev       & Test       & Train       & Dev        & Test      \\ \hline \hline
        Full           & 2013        & 252       & 251        & 19940       & 2493       & 2491      \\
        Partial        & 420         & 252       & 251        & 3970        & 2493       & 2491      \\ \hline
        \end{tabular}
    }
    \caption{Size of the different splits for the benchmarking of each architecture.}
    \label{tab:sample-size}
\end{table}

\section{Experiments}
\label{sec:experiments}

\subsection{Baselines}

We designed four baselines with the objective of evaluating whether or not we were outperforming a full-text approach in our sentence classification task. These baselines leverage the annotations we created for each lemma associated with sexual semantics in our dataset. Each baseline operates by classifying a sample sentence as positive if it contains any of the known lemmas from our annotations.

\begin{description}
    \item[Baseline 1:] This baseline includes the complete list of lemmas from our dataset, totaling 702 lemmas.
    \item[Baseline 2:] Similar to Baseline 1, but stopwords are excluded based on a stopword list from \citetlanguageresource{aurelien_berra_2020_3860343}, resulting in 684 lemmas.
    \item[Baseline 3:] This baseline utilizes the list of lemmas from Baseline 2 but further filters out sexual phrases consisting of two or more words, resulting in 537 lemmas.
    \item[Baseline 4:] Building on Baseline 3, this baseline excludes lemmas related to sexual phrases that are based on metonymy or metaphors, leaving 218 lemmas.
\end{description}

% In essence, these baselines progressively refine the set of lemmas used for classification, with the last baseline being the most restrictive, focusing on more specific instances of sexual content detection.

\subsection{Experimental setup}
\label{sub:exp-setup}

Model were trained using AllenNLP~\citep{Gardner_AllenNLP_A_Deep}, with following hyper-parameters: batch size of 4, learning rate of 1e-4, patience of 5 and maximum epochs of 50\footnote{These parameters were found to be the best and the most stable across all configurations, through a parameter optimization search using Optuna\cite{akiba2019optuna}. The search space for batch-size ranged from 4 to 64, the learning rate from 1e-2 to 1e-5.}. 

BERT pre-trained weights are using Latin BERT, other pretrained embeddings are trained with the same search corpus, using lemma or lexemes. All embeddings except characters were frozen during training. 

Regarding BERT, we tested both pooling at the last and penultimate layers, the last layer yield consistently better or similar results. For the partial corpus experiment, only MeanMax is shown in terms of BERT-pooling, as other would not converge. We had a similar issue with fine-tuning BERT or RoBERTa models accross datasets.

RNNs (LSTM, HAN, GRU) are trained with two different dimension (64 per direction and 128), only the best are kept for scores (dimension is shown).

Each architecture is trained 10 times with random seeds, and the average and standard deviation is used for evaluation. Model are ranked according to their precision, as it indicates how much noise a human would have to filter in order to expand their corpus, but provides also true positive rate~(TPR, $tp/(tp+fn)$), true negative rate~(TNR, $tn/(fp+tn)$), and F1-Score.

\subsection{Sub-experiment: Testing bias in the context of categorical features}
\label{sub:testing-bias}

In order to evaluate the impact and potential bias of integrating categorical features, we conducted a secondary experiment. In this experiment, we selected a well-performing architecture and tested it with different combinations of categorical features: (a) using all of them, (b) using all except the author identification, (c) using only the form (prose/verse) and the century, and (d) using none of the categorical features.

The objective of configuration (b) was to retain less specific information about the author while maintaining detailed information about the text structure. Configuration (c) focused solely on the sociolect by including very generic information.

To assess the effectiveness of this approach, we selected two complete texts and "disguised" the text as if it belonged to another author. The texts chosen were Cicero's \textit{De finibus}, a philosophical dialogue with normally no or very minimal sexual content, and Martial's \textit{Epigrammata}. For each text, we annotated all their sentences using their correct metadata and the metadata of their disguised version.

\section{Results}
\label{sec:results}

\subsection{Qualitative evaluation of categorical feature use}

In the preliminary results, when using categorical features in a similar way to \citet{kim_categorical_2019}, we observed significant differences in scores between models that incorporated these features and models that ignored any extra-textual information. The precision and true positive rate, which are inherently the metrics we want to maximize, reached up to 89.33\% (+3 points) and 80.08\% (+10 points). However, when conducting the experiment described in subsection \ref{sub:testing-bias}, we observed a significant overfitting phenomenon specifically related to categorical features (\textit{cf.} Table \ref{tab:martial-cicero}).

This overfitting phenomenon is likely attributed to the substantial bias present in the dataset towards certain authors, particularly Martial, who is over-represented in relation to his weight in the overall Latin corpus, and conversely, Cicero, who is under-represented. The stark contrast in scores between Martial and Cicero (Martial being ten times more sexual when seen as himself compared to being seen as Cicero when using all metadata, while Cicero being 40 times more likely to talk about sexuality when seen as Martial compared to being seen as himself when using Form and Century metadata) led us to unfortunately exclude categorical features as an option for our biased corpus.

\begin{table}[ht]
    \centering
    \resizebox{\linewidth}{!}{%
    \begin{tabular}{llrr}
    \hline
                   Text & Categorical features        & As Cicero's & As Martial's \\
    \hline \hline
    \textit{De Finibus}, Cicero & All                       &               4,25\% &              72,60\% \\
    \textit{Epigrammata}, Martial &                              &               7,89\% &              82,05\% \\ \hline
    \textit{De Finibus}, Cicero & All but authors           &               4,69\% &              47,54\% \\
    \textit{Epigrammata}, Martial &                              &               7,33\% &              59,76\% \\ \hline
    \textit{De Finibus}, Cicero & Form and Century              &               0,91\% &              36,03\% \\
    \textit{Epigrammata}, Martial &                              &               6,41\% &              48,10\% \\ \hline
    \textit{De Finibus}, Cicero & None                       &               2,91\% &               2,91\% \\
    \textit{Epigrammata}, Martial &                              &              15,29\% &              15,29\% \\ \hline
    \end{tabular}%
    }
    \caption{Positively tagged sentences for sexual content, depending on the categorical features used and the metadata provided. Logically, there are no differences when no categorical feature is used.}
    \label{tab:martial-cicero}
\end{table}

\subsection{Architecture benchmarking}

\paragraph{Main split}

Given our objectives of providing a novel solution for easily expanding or creating datasets around specific semantics in the fields of digital humanities and historical linguistics, our approach yields robust results despite the limited size of our positive training samples (\textit{cf.} Table \ref{tab:main-scores}). Considering both precision and True Positive Rate (TPR) metrics, we find that the HAN-256 model outperforms other approaches in alignment with our objectives. It achieves a commendable TPR of 70\% along with a reasonably high precision of 85\%. Striking a balance between missing 30\% of positive examples while only requiring the filtration of 15\% of noise, this approach appears to be a favorable compromise, surpassing all baseline methods that rely on token matching.

As for the input data used, it is noteworthy that BERT embeddings and Token embeddings exhibit significantly inferior performance compared to the RNN-based approach utilizing lemma embeddings. The stark contrast between token and lemma embeddings is particularly striking, especially when examining TPR, where there is a notable difference of -13 points between the HAN models. This disparity leads us to hypothesize that lemmatization effectively mitigates noise in our dataset, negating the effects of the rich morphology of Latin -- a challenge that character embeddings alone do not seem to be able to adequately address. The situation with BERT embeddings presents a more intricate analysis, and we conjecture that the dimensionality of the embeddings may pose challenges given the inherent imbalance and scale of our training dataset.

Despite having the lowest precision among all the pooling methods, MeanMax stands out with the highest True Positive Rate (TPR) among them all. This remarkable disparity in TPR, reaching up to +24 points when compared to the Max pooling method, coupled with a relatively modest difference in precision (-2 points compared to Max), leads us to designate MeanMax as the most favorable pooling method in this context.

\begin{table}[ht]
\resizebox{\linewidth}{!}{%}
\begin{tabular}{llllll}
\hline
{} & & TPR & TNR &   Precision &          F1 \\
Embedding & Model             &                   &                   &             &             \\ \hline
 & Baseline 1 & 100.0 & 7.43 & 9.82 & 17.88  \\
 & Baseline 2 & 100.0 & 12.32 & 10.31 & 18.69  \\
 & Baseline 3 & 98.8 & 48.37 & 16.17 & 27.79  \\
 & Baseline 4 & 74.9 & 72.34 & 21.44 & 33.34  \\
\hline
Lemma & GRU-256           &        65.78 ± 5.82 &        \textbf{98.92 ± 0.64} &  \textbf{86.65 ± 5.13} &  74.45 ± 2.59 \\
Lemma & HAN-256           &        \textbf{70.60 ± 2.83} &        98.76 ± 0.30 &  85.33 ± 2.75 &  \textbf{77.20 ± 1.35} \\
Token & HAN-128           &        57.81 ± 5.37 &        98.82 ± 0.42 &  83.51 ± 3.79 &  68.06 ± 3.11 \\
Token & GRU-128           &        59.16 ± 5.37 &        98.74 ± 0.35 &  82.81 ± 3.37 &  68.81 ± 3.16 \\
Token & LSTM-128          &        57.01 ± 3.11 &        98.76 ± 0.26 &  82.33 ± 2.52 &  67.29 ± 1.84 \\
Lemma & LSTM-256          &        66.73 ± 5.07 &        98.47 ± 0.67 &  82.16 ± 5.89 &  73.32 ± 2.26 \\
BERT  & GRU-256           &        66.97 ± 5.36 &        98.17 ± 0.33 &  78.79 ± 2.14 &  72.25 ± 2.91 \\
BERT  & HAN-256           &        68.39 ± 4.61 &        98.02 ± 0.39 &  77.91 ± 2.44 &  72.69 ± 1.98 \\
BERT  & Mean              &        50.08 ± 3.33 &        98.29 ± 0.12 &  74.73 ± 0.64 &  59.91 ± 2.45 \\
BERT  & Max               &        30.52 ± 6.64 &        98.86 ± 0.39 &  73.41 ± 3.08 &  42.62 ± 6.32 \\
BERT  & [BOS]               &        44.98 ± 4.20 &        98.22 ± 0.32 &  71.92 ± 2.19 &  55.20 ± 2.90 \\
BERT  & MeanMax           &        54.18 ± 5.52 &        97.76 ± 0.59 &  71.34 ± 3.18 &  61.31 ± 2.76 \\
\hline
\end{tabular}
}
\caption{Median scores over 10 trained models for each model per embedding and encoder. ``Lemma'' or ``token'' stand for the concatenation of word level embedding and character level embeddings of the given input.}
\label{tab:main-scores}
\end{table}

\paragraph{Partial split}
When applied to the same testing set but with a considerably smaller dataset, we observe substantial shifts in the performance landscape (\textit{cf.} Table \ref{tab:partial-scores}). BERT, employing MeanMax encoding, emerges as the leading model in terms of balance between precision and TPR (Token based input have a very low TPR). Conversely, RNN, utilizing BERT embeddings as input, outperforms MeanMax by nearly 14 points in True Positive Rate (TPR), with only a marginal 1-point decrease in precision. Among the non-BERT-based models, lemma embeddings combined with the HAN model consistently rank in the top three, exhibiting competitive precision and TPR scores. While precision scores remain relatively consistent within the dataset, TPR scores reveal a noticeable 7-point difference between the top-ranked and second-ranked models.

In comparison to models trained on larger training samples, these smaller datasets yield significantly lower scores, with a decrease of -15 in precision and -7 in TPR. Nonetheless, these scores continue to outperform the token search baseline by a substantial margin. It is worth noting that human filtering of the model's output remains a viable option given the achieved precision, which entails filtering out approximately 30\% of noise from the results.

\begin{table}[ht]
    \centering
    \resizebox{\linewidth}{!}{%}
        \begin{tabular}{llllll}
        \hline
         & {} & TPR & TNR &   Precision &          F1 \\
        Embedding & Model            &                   &                   &             &             \\
        \hline
         &Baseline 1 & 100.0 & 7.43 & 9.82 & 17.88  \\
         &Baseline 2 & 100.0 & 12.32 & 10.31 & 18.69  \\
         &Baseline 3 & 98.8 & 48.37 & 16.17 & 27.79  \\
         &Baseline 4 & 74.9 & 72.34 & 21.44 & 33.34  \\
        \hline
        Token & GRU-128    &        22.99 ± 3.92 &        99.27 ± 0.40 &   77.95 ± 8.74 &  35.09 ± 4.08 \\
        Token & LSTM-128   &        27.37 ± 4.78 &        98.92 ± 0.69 &   74.00 ± 8.29 &  39.41 ± 3.78 \\
        BERT  & MeanMax    &        49.80 ± 3.34 &        \textbf{98.02} ± 0.24 &  \textbf{71.74} ± 1.71 &  58.71 ± 2.27 \\
        Token & HAN-256    &        23.63 ± 6.78 &        98.96 ± 0.46 &   71.56 ± 7.70 &  34.67 ± 8.38 \\
        BERT  & HAN-256    &        \textbf{63.11} ± 3.79 &        97.30 ± 0.63 &  70.54 ± 3.60 &  \textbf{66.43 ± }0.80 \\
        Lemma & HAN-256    &        51.51 ± 5.93 &        97.70 ± 0.67 &  69.77 ± 5.16 &  58.95 ± 3.61 \\
        BERT  & GRU-256    &        56.61 ± 7.71 &        97.25 ± 1.03 &  68.36 ± 4.78 &  61.39 ± 2.94 \\
        Lemma & LSTM-256   &        47.37 ± 8.79 &        97.38 ± 1.46 &  66.43 ± 7.32 &  54.35 ± 3.67 \\
        Lemma & GRU-128    &        48.92 ± 6.03 &        97.33 ± 0.95 &  65.74 ± 5.41 &  55.64 ± 2.95 \\
        \hline
        \end{tabular}
    }
    \caption{Median scores over 10 trained models for each model per embedding and encoder, on the partial dataset. ``Lemma'' or ``token'' stand for the concatenation of word level embedding and character level embeddings of the given input.}
    \label{tab:partial-scores}
\end{table}

\paragraph{HAN as a human-centered solution}

With HAN models consistently ranking at the top of our models, we seized the opportunity to examine the relationship between attention weights at the token level and the identified words bearing sexual semantics. To investigate this, we calculated the relative rank of ``sexual'' tokens (according to the original analysis of Adams) by dividing their attention rank by the total number of tokens in the sentence. For instance, in a ten-word sentence, the first token would have a relative rank of 10\%.

\begin{figure}[ht]
    \centering
    \includegraphics[width=\linewidth]{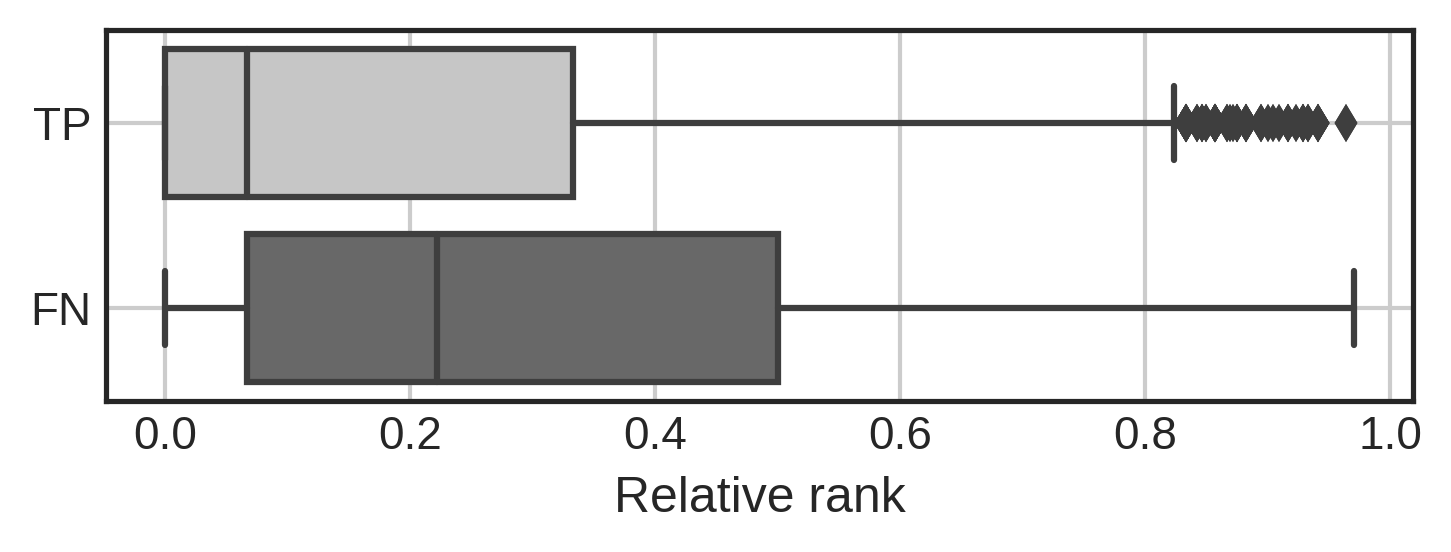}
    \caption{Relative rank distribution of the sexual terms in TP and FN.}
    \label{fig:attention}
\end{figure}

The results (Fig. \ref{fig:attention}) demonstrate the model's ability to identify the salient semantic features of the sentence. This is a significant finding, as it enables us to leverage attention weights to assist human readers in their corpus filtering task, directing the philologist's attention towards decision-making factors.

Additionally, we discovered that in negative samples, punctuation significantly attract attention: 27\% of exclamation marks, 35.6\% of dots, 22.3\% of semicolons, and 48.3\% of question marks have the highest attention weights of their sentences% (cf. Table \ref{tab:punc})
. We expect that the reason behind this fallback to punctuation for attention in negative samples is that punctuation is at least common to all negative samples and positive samples, thus having no or nearly no semantic value.

\section{Conclusion}

In this paper, we presented a comprehensive analysis and evaluation of various architectures for a sentence semantic classification task in Latin. Our study focused specifically on addressing the challenges of the diversity of sexual idioms, subjects and genres.

Throughout our research, we thoroughly investigated various approaches to embedding and encoding, taking into account linguistic and contextual information in our models. We consistently observed strong performance from models utilizing LSTM, GRU, and HAN encoders, highlighting their efficacy in capturing the intricacies of the Latin language, particularly when using lemmas. We also showed that, while the inclusion of categorical features, such as author identity, yielded score improvements, these enhancements were inherently linked to biases present within the corpus.

Moreover, our experiments underscore the substantial impact of dataset size on model performance. Larger datasets consistently yielded higher scores, whereas smaller datasets exhibited a pronounced decline in performance across multiple metrics. Furthermore, our findings reveal that the incorporation of BERT did not bring about a significant alteration in overall performance and, in some cases, even underperformed in the context of the larger dataset experiment. This suggests that, especially in compute-limited scenarios or for languages lacking MLM, the performance of RNNs remains promising, and their effectiveness should not be underestimated, even when considering the performance of BERT models.

These results further underscore the efficacy of the HAN model in capturing and attending to pertinent semantic features, offering valuable insights into the interpretability of the model's attention mechanism. The discernment of key attention weights, as well as the significance of punctuation marks in negative samples, enhances our comprehension of the model's behavior. This understanding can play a pivotal role in refining the computational analysis of sexual content in ancient texts. The attention scores generated by the model hold immense significance, particularly in assisting researchers in filtering out false positives when examining newly tagged data.

Although our study provides valuable insights and performance comparisons across various models and techniques, it also highlights areas ripe for future research. Notably, the underperformance of lexemes and character-level embeddings for lexemes presents an intriguing avenue for further investigation. Exploring features that the model may routinely overlook for classification purposes and verifying whether our hypothesis regarding the impact of morphological richness on the score drop is accurate could shed light on these underperforming aspects. The conclusions drawn in this paper should also undergo testing in various other semantic fields, including those with lower levels of lexical inventiveness. This broader exploration can help us identify the potential limitations of applying similar techniques and models in different contexts.

Overall, our findings contribute to the understanding of Latin language processing within the computational linguistic community, paving the way for future advancements in this field. By leveraging innovative approaches and carefully considering the nuances of the Latin language, we can unlock new possibilities for automated analysis, interpretation, and understanding of Latin texts, ultimately providing valuable assistance to philologists in their research endeavors.

\section*{Limitations}

While our study provides valuable insights into semantic classification in Latin, there are several limitations to consider:

\begin{enumerate}
    \item Our focus on the sexual semantic field may limit the generalizability of our findings to other domains, as different semantic fields may have unique linguistic characteristics that require tailored approaches.
    \item Although we incorporated morpho-syntactic features as embeddings, there may exist other embedding sizes where they could have performed better as individual features rather than as agglomerated ones.
    \item Our evaluation of sentence encoders was not exhaustive, and exploring alternative architectures, such as fine-tuned BERT, could yield different results. However, it should be noted that not every language has access to or the ability to build a masked language model.
    \item The specific bias introduced by choosing the sexual semantic field as our evaluation subject could influence the performance of categorical features, leading to potential limitations in their generalizability.
\end{enumerate}

Addressing these limitations and expanding the scope of future studies will enhance our understanding and application of deep learning methods in Latin language processing.

\section*{Ethics Statement}

Reader should be warned that Roman literature, being a product of its time, can contain explicit and potentially disturbing content related to sex and sexuality. It is important to acknowledge that the cultural and societal norms of the ancient Roman era as well as of High Middle Ages differ significantly from contemporary values and attitudes. The original dataset includes instances of explicit language, depictions of sexual acts, and even instances of humor involving rape threats.

While we recognize the importance of ethical considerations in computational linguistics, including the responsible treatment of sensitive subjects, the substantial temporal gap between the analyzed texts and the present day provides an additional layer of separation that helps mitigate immediate ethical concerns regarding the content of the dataset in terms -- but not exclusively -- of privacy. However, it is crucial to remain vigilant and address any potential biases or ethical implications that may arise from the interpretation and analysis of historical texts.

\nocite{*}
\section{Code and data availability statement}

Code (under MPL License) and training dataset (CC-BY 4.0) are available at \url{https://github.com/lascivaroma/seligator}. The positive samples can be found in their original repository at \url{https://github.com/lascivaroma/exemplier}.

% Links to repositories and code will be provided after the review, under open licences (CC-BY for corpora, MPL for code).

\section{Bibliographical References}\label{reference}
%\label{main:ref}

\bibliographystyle{lrec_natbib}
\bibliography{bibliography}

\section{Language Resource References}
\label{lr:ref}
\bibliographystylelanguageresource{lrec_natbib}
\bibliographylanguageresource{languageresource}

\end{document}